%% file: main.tex
\documentclass{article}
\usepackage{ecod_preprint,times}
\usepackage[T1]{fontenc}
\usepackage[utf8]{inputenc}
\usepackage{amsmath,amssymb,graphicx,booktabs,xcolor}
\usepackage{longtable,array,calc,multirow}
\usepackage{algorithm}
\usepackage{float,needspace,wrapfig}
\usepackage{algpseudocode}
\usepackage[normalem]{ulem}
\usepackage{hyperref,url}
\input{authors}
\hypersetup{colorlinks=true,citecolor=blue,linkcolor=blue,urlcolor=blue,
  pdftitle={ECOD: Experience-Calibrated On-Policy Distillation for Multi-Turn Agents},
  pdfauthor={\ECODPDFAuthors}}

\newcommand{\deleted}[1]{}
\providecommand{\tightlist}{\setlength{\itemsep}{0pt}\setlength{\parskip}{0pt}}
\title{ECOD: Experience-Calibrated On-Policy\\Distillation for Multi-Turn Agents}
\author{\ECODAuthorBlock}
\date{}
\begin{document}
\maketitle
\begin{abstract}
\input{sections/abstract}
\end{abstract}
\section{Introduction}\label{sec:intro}
\input{sections/introduction}
\section{Related Work}\label{sec:related}
\input{sections/related_work}
\section{Preliminaries}\label{sec:preliminaries}
\input{sections/preliminaries}
\section{Experience-Calibrated On-Policy Distillation}\label{sec:method}
\input{sections/method}
\section{Experiments}\label{sec:experiments}
\input{sections/experiments}
\section{Conclusion}\label{sec:conclusion}
\input{sections/conclusion}
\label{main-text-end}
\section*{AI Use Statement}
\input{sections/ai_use_statement}
\bibliography{references}
\bibliographystyle{iclr2027_conference}
\clearpage
\appendix
\input{sections/appendix}
\end{document}

%% file: authors.tex
\newcommand{\ECODPDFAuthors}{Jie Jin, Shengqin Wang, Yan Gao, Xin Tan, Yuan Xie}
\newcommand{\ECODAuthorBlock}{%
  \textbf{Jie Jin\thanks{Equal contribution.}\kern0.3em, Shengqin Wang\textsuperscript{*},
  Yan Gao, Xin Tan, Yuan Xie}\\
  \parbox[t]{\dimexpr\textwidth-2\tabcolsep\relax}{%
    \normalfont\raggedright
    \mbox{East China Normal University};
    \mbox{Shanghai Innovation Institute};
    \mbox{Shanghai Artificial Intelligence Laboratory}}}

%% file: sections/abstract.tex
Sparse and delayed rewards make reinforcement learning challenging for long-horizon language agents. On-policy distillation (OPD) complements environmental rewards with dense teacher feedback on student-generated contexts. Standard OPD can misalign supervision with relative teacher competence, student trajectory outcomes, and the temporal structure of task execution. Recent variants adjust guidance through divergence-based weighting and temporal schedules, while joint calibration from task-completion experience remains underexplored. We introduce Experience-Calibrated On-Policy Distillation (ECOD), which allocates teacher supervision at the task, trajectory, and turn levels. ECOD compares teacher and student success-rate posteriors under matched task conditions, restricts direct supervision to failed student trajectories, and derives a temporal prior from successful teacher lengths. The resulting weight applies to teacher losses or teacher-derived advantages. We evaluate ECOD on ALFWorld and WebShop with Qwen2.5-1.5B-Instruct and Qwen2.5-7B-Instruct. Under matched teacher settings, ECOD improves standard OPD by up to 4.06\% in relative ALFWorld unseen success. Across both model scales, ECOD-assisted SOD, TCOD-F2B, and ATOD configurations yield average paired relative gains of 6.20\% on ALFWorld unseen and 7.05\% on WebShop, with WebShop gains ranging from 2.29\% to 21.60\%. These results support experience-calibrated supervision allocation as a reusable mechanism across distinct distillation objectives.

%% file: sections/introduction.tex
Long-horizon language agents must coordinate sequential decisions under partial observability, often with sparse or delayed environmental rewards \citep{react}. This makes credit assignment across extended interaction trajectories a central challenge for reinforcement learning (RL). Methods such as GiGPO improve multi-turn policy optimization, but sparse rewards still provide limited feedback on individual decisions \citep{gigpo}.

On-policy distillation (OPD) complements sparse environmental rewards with dense supervision from a fixed teacher evaluated on student-generated contexts \citep{gkd}. Standard OPD can suffer from teacher supervision misalignment when guidance overlooks the teacher's relative competence, student trajectory outcomes, or the temporal structure of task execution. Recent variants make teacher guidance more adaptive through divergence-based weighting, temporal curricula, and turn-aware scheduling \citep{sod,tcod,atod}. These methods improve specific aspects of supervision, but their adaptation remains tied to individual signals or mechanisms. A broader question is how teacher supervision should be allocated when multiple aspects of the interaction jointly determine whether guidance is useful.

\begin{figure}[t]
\centering
\setlength{\abovecaptionskip}{5pt}
\includegraphics[width=\linewidth]{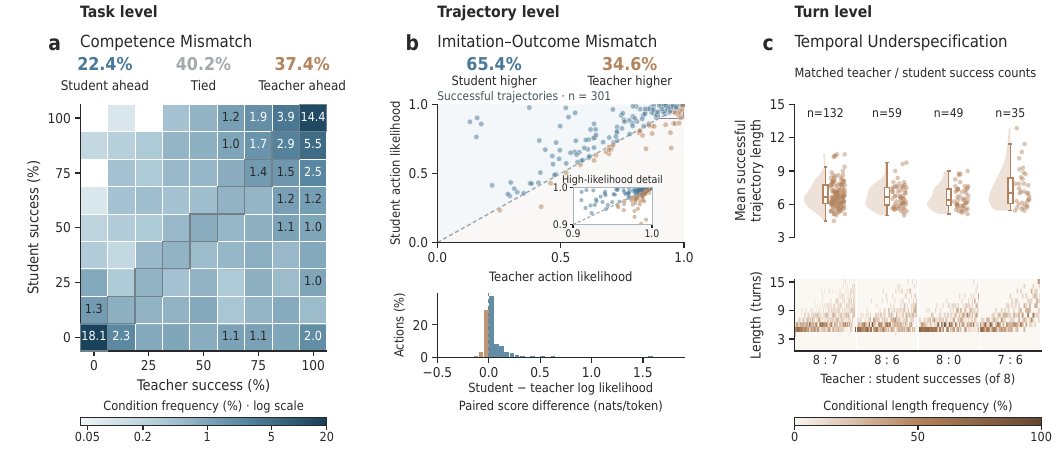}
\caption{\small \textbf{Empirical motivation for experience-calibrated teacher supervision across task, trajectory, and turn levels in synthetic WebShop (1,000 products).} (a) Task level: matched teacher--student success rates. (b) Trajectory level: conditional action scores on successful student trajectories. (c) Turn level: successful teacher trajectory lengths grouped by matched success counts.}
\label{fig:motivation}
\end{figure}

We examine this question through three recurring patterns in teacher--student interactions on synthetic WebShop (Figure~\ref{fig:motivation}). At the task level, the teacher records no more successes than the student in 62.6\% of 2,400 matched task-condition observations with eight rollouts per policy (Figure~\ref{fig:motivation}(a)). This Competence Mismatch motivates estimating the teacher's local completion advantage. At the trajectory level, 197 of 301 actions sampled from successful student trajectories receive higher student than teacher conditional scores under identical histories (Figure~\ref{fig:motivation}(b)). This Imitation--Outcome Mismatch shows that successful task completion does not imply agreement in action likelihood. At the turn level, conditions with eight teacher successes and seven student successes have mean successful teacher lengths ranging from 4.5 to 10.5 turns (Figure~\ref{fig:motivation}(c)). This Temporal Underspecification shows that completion counts alone do not capture execution length. These observations motivate supervision weights informed by relative competence, individual outcomes, and temporal completion evidence. Appendix~\ref{sec:motivation-protocol} describes the data sources, sampling, and scoring protocol.

\begin{wrapfigure}{r}{0.52\textwidth}
\vspace{-10pt}
\centering
\setlength{\abovecaptionskip}{3pt}
\includegraphics[width=\linewidth]{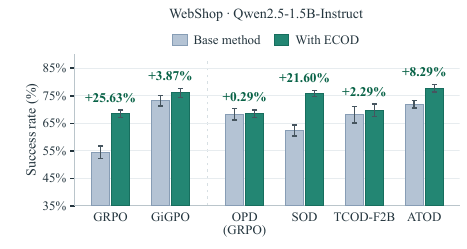}
\caption{\small \textbf{ECOD on WebShop.} Means ± SD over eight decoding runs; labels show relative gains. GRPO and OPD share the ECOD result.}
\label{fig:intro-results}
\vspace{-8pt}
\end{wrapfigure}

To integrate these three sources of evidence, we propose \textbf{Experience-Calibrated On-Policy Distillation (ECOD)}, a unified mechanism for allocating teacher supervision across tasks, trajectories, and turns. Local competence calibration determines task-level supervision strength by comparing teacher and student success-rate posteriors under matched task conditions. Outcome-conditioned selection concentrates direct teacher supervision on failed student trajectories, while all trajectories remain available for reward-based learning. A temporal supervision prior uses successful teacher execution lengths to distribute guidance across turns. These three components form a bounded supervision weight applicable to teacher losses or teacher-derived advantages, allowing the same allocation rule to be integrated into different OPD objectives. Inference uses only the student.

We evaluate ECOD on ALFWorld and WebShop with Qwen2.5-1.5B-Instruct and Qwen2.5-7B-Instruct, covering GRPO, GiGPO, standard OPD, and three OPD variants (Tables~\ref{tab:main-15b} and~\ref{tab:main-7b}). The reported checkpoint comparisons with the 1.5B student yield relative ALFWorld unseen success gains of 4.06\% and 2.56\% over GRPO-based and GiGPO-based OPD, respectively, under matched teacher settings. Across both model scales, ECOD-assisted SOD, TCOD-F2B, and ATOD configurations average paired relative gains of 6.20\% on ALFWorld unseen and 7.05\% on WebShop. Figure~\ref{fig:intro-results} summarizes the 1.5B WebShop results. Fixed-endpoint component, teacher-strength, and uniform-allocation controls evaluate supervision allocation separately from the main checkpoint comparisons (Figure~\ref{fig:ablation}).

Our contributions are:

\begin{itemize}
\tightlist
\item We characterize teacher supervision misalignment in multi-turn OPD at the task, trajectory, and turn levels, and provide empirical evidence motivating unified supervision allocation across these levels.
\item We propose ECOD, which jointly calibrates task-level strength, trajectory selection, and turn-level allocation through a bounded supervision weight applicable to teacher-loss and teacher-advantage formulations.
\item We evaluate ECOD across RL backbones and OPD variants on ALFWorld and WebShop, and assess the contribution of selective supervision allocation through component ablations and teacher-strength controls. 
\end{itemize}

%% file: sections/related_work.tex
\hypertarget{related_work-long-horizon-agent-reinforcement-learning}{%
\subsection{Long-horizon agent reinforcement learning}\label{related_work-long-horizon-agent-reinforcement-learning}}

Research on language agents covers reasoning and feedback during interaction \citep{react,reflexion}, instruction tuning, and environment-based policy learning \citep{agenttuning,agentgym}. WebRL~\citep{webrl} uses online curricula, AgentGym-RL~\citep{agentgymrl} expands interaction horizons, and GiGPO~\citep{gigpo} improves step-level credit assignment. ECOD adds teacher supervision to reward-based training and computes its weights from completion outcomes and successful teacher execution lengths.

\hypertarget{related_work-on-policy-distillation}{%
\subsection{On-policy distillation}\label{related_work-on-policy-distillation}}

Sequence-level distillation uses teacher-generated outputs \citep{seqkd}, while DAgger~\citep{dagger} and OPD~\citep{gkd} obtain supervision on learner-generated contexts. MiniLLM studies reverse-KL teacher losses \citep{minillm}. Agentic methods also incorporate teacher feedback through policy advantages \citep{atod,safopd}. Teacher guidance can be adjusted using divergence, uncertainty, annealing, relative completion rates, and token or trajectory gates \citep{sod,atod,retireopd,sdar,seedopd,serl}.

Methods also differ in where they intervene. TCOD and TurnOPD control interaction depth \citep{tcod,turnopd}. Guided-OPD mixes teacher and student turns \citep{guidedopd}, while ReOPD and STRIDE select replay or restart positions \citep{reopd,stride}. FutureBridge-OPD and OVCSD assess guidance through future continuations or outcomes \citep{futurebridge,ovcsd}. OPID, MOPD, and CCOPD construct skill-, peer-, and context-based targets \citep{opid,mopd,ccopd}.

ECOD computes an allocation weight from task-matched completion outcomes and successful teacher lengths. It applies this weight to the teacher contribution while retaining the base method's reward-advantage construction and online student rollout procedure. The allocation rule supports explicit teacher losses and teacher-derived advantages. Appendix~\ref{sec:recent-opd-comparison} compares the evidence, update terms, and rollout operations of related methods.

%% file: sections/preliminaries.tex
\hypertarget{preliminaries-reinforcement-learning-for-language-agents}{%
\subsection{Reinforcement learning for language agents}\label{preliminaries-reinforcement-learning-for-language-agents}}

A language agent \(\pi_\theta\), parameterized by \(\theta\), observes history \(h_t\) and generates a response \(x_t\) containing reasoning and an executable action at turn \(t\). A trajectory is \(\tau=(h_t,x_t)_{t=0}^{L-1}\), where \(L\) is its length in action turns. Reinforcement learning maximizes expected environment return using reward-derived advantages to update the policy \citep{ppo}. For a task condition \(u\), GRPO samples \(G\) trajectories from the rollout policy \(\pi_{\mathrm{old}}\) and computes group-relative advantages without a learned critic \citep{grpo}. Its clipped policy objective is

\begin{equation}
\mathcal J_{\mathrm{GRPO}}^{\mathrm{clip}}(\theta)
=\mathbb E\!\left[
\min\!\left(
r_{k,j}\widehat A_k,\,
\operatorname{clip}(r_{k,j},1-\epsilon,1+\epsilon)\widehat A_k
\right)
\right].
\label{eq:grpo}
\end{equation}

The expectation averages over task conditions and sampled groups, uniformly over trajectories and then response tokens within each trajectory. Here, \(r_{k,j}=\pi_\theta(y_{k,j}\mid c_{k,j})/\pi_{\mathrm{old}}(y_{k,j}\mid c_{k,j})\) is the likelihood ratio for response token \(y_{k,j}\) given history and response prefix \(c_{k,j}\). The advantage is \(\widehat A_k=(R_k-\bar R)/(\sigma_R+\delta)\), where \(R_k\) is the trajectory return and \(\bar R,\sigma_R\) are its group mean and standard deviation. The clipping radius is \(\epsilon\), and \(\delta>0\) ensures numerical stability. We denote the negative policy objective together with the base algorithm's regularizers by \(\mathcal L_{\mathrm{RL}}\).

\hypertarget{preliminaries-on-policy-distillation}{%
\subsection{On-policy distillation}\label{preliminaries-on-policy-distillation}}

On-policy distillation (OPD) evaluates a fixed teacher \(\pi_T\) on student-generated contexts, reducing the distribution mismatch associated with fixed demonstrations \citep{gkd}. Teacher and student predictions condition on the same interaction history and response prefix. With an explicit distillation loss \(\mathcal L_{\mathrm{distill}}\), the combined training objective is

\begin{equation}
\mathcal L(\theta)
=\mathcal L_{\mathrm{RL}}(\theta)
+\lambda\,\mathcal L_{\mathrm{distill}}(\theta),
\qquad \lambda\geq0.
\label{eq:opd}
\end{equation}

Here, \(\lambda\) controls teacher supervision strength. Alternatively, teacher feedback can enter the policy update through a teacher-derived advantage. ECOD's weighting of these two forms is defined in Section~\ref{sec:method}.

%% file: sections/method.tex
\begin{figure}[!ht]
\centering
\setlength{\abovecaptionskip}{5pt}
\includegraphics[width=\linewidth]{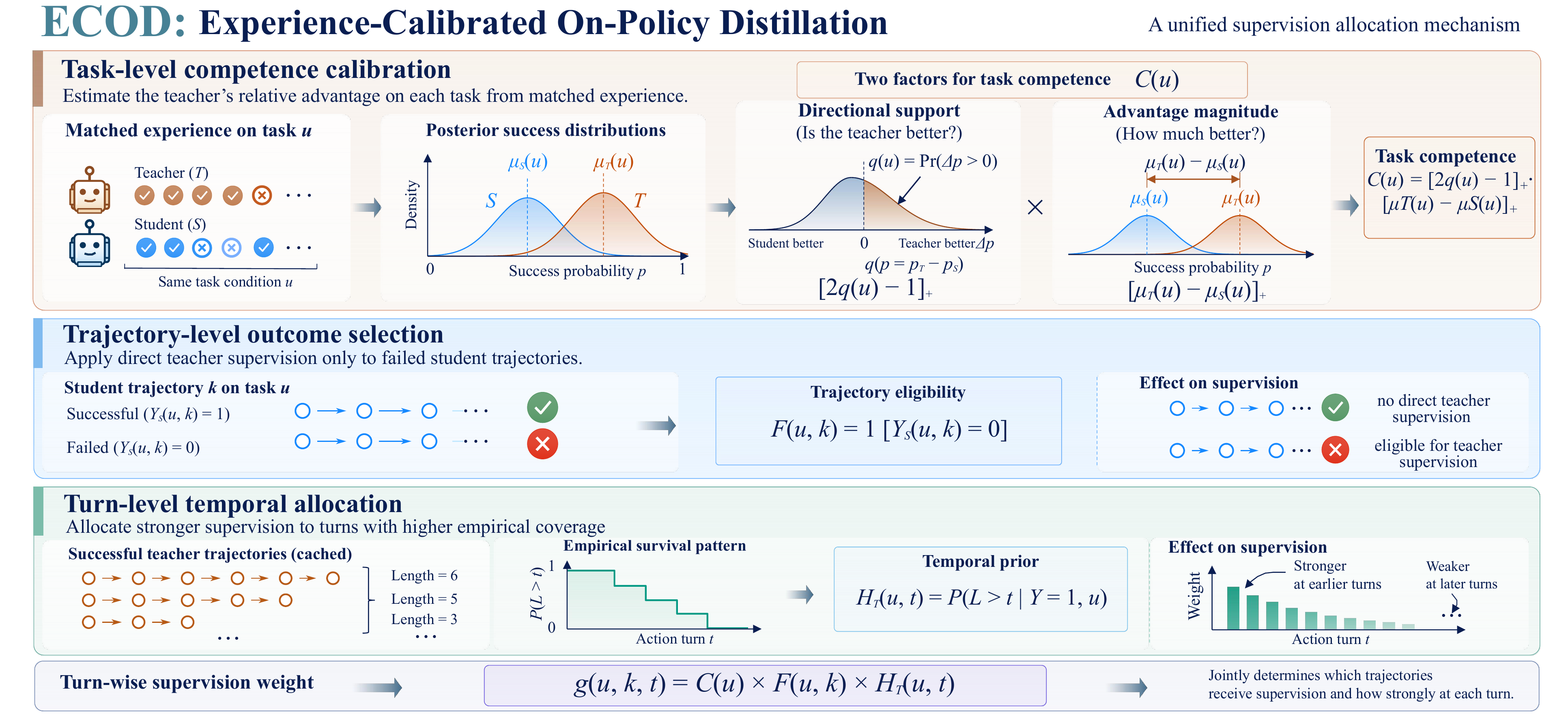}
\caption{\small \textbf{The framework of ECOD.} Local competence calibration, outcome-conditioned selection, and a temporal supervision prior jointly determine teacher supervision across tasks, trajectories, and turns.}
\label{fig:method}
\end{figure}

ECOD computes teacher-supervision weights from cached teacher rollouts and the current student rollout batch (Figure~\ref{fig:method}). For each task condition, it compares teacher and student completion outcomes, selects failed student trajectories, and assigns turn weights from successful teacher lengths. These weights are held fixed during optimization and applied to the teacher contribution.

\hypertarget{method-experience-calibrated-supervision}{%
\subsection{Experience-calibrated supervision}\label{method-experience-calibrated-supervision}}

Let \(u\) denote a task and its initial environment configuration, \(k\) a student trajectory, and \(t\) its zero-based action-turn index. ECOD assigns each turn a supervision weight:

\begin{equation}
g(u,k,t)=
\underbrace{C(u)}_{\text{local competence}}\times
\underbrace{F(u,k)}_{\text{outcome selection}}\times
\underbrace{H_T(u,t)}_{\text{temporal prior}} .
\label{eq:ecod-weight}
\end{equation}

Here, \(C(u)\) controls task-level strength, \(F(u,k)\) selects failed trajectories, and \(H_T(u,t)\) allocates supervision across turns. Their joint effect is illustrated at the bottom of Figure~\ref{fig:method}.

\hypertarget{method-local-competence-calibration}{%
\subsubsection{Task-level competence calibration}\label{method-local-competence-calibration}}

Local competence calibration estimates the teacher's completion advantage from task-matched rollout outcomes (Figure~\ref{fig:method}, first row).

For task \(u\), records comprise \(n_T(u)\) cached teacher trajectories, \(\mathcal D_T(u)=\{(Y_{T,u,i},L_{T,u,i})\}_{i=1}^{n_T(u)}\), and \(n_S(u)\) current student trajectories, \(\mathcal D_S(u)=\{Y_{S,u,k}\}_{k=1}^{n_S(u)}\). Here, \(i\) indexes teacher trajectories, \(Y\in\{0,1\}\) records failure/success, and \(L\) counts action turns. Student statistics use only the current rollout policy \(\pi_{\mathrm{old}}\).

For teacher or student \(a\in\{T,S\}\), let \(p_a\) be its task-conditioned completion probability under its collection protocol. With conditionally independent Bernoulli outcomes and independent \(\operatorname{Beta}(1,1)\) priors, \(s_a=s_a(u)\) successes in \(n_a=n_a(u)\) rollouts give the posterior and mean

\begin{equation}
p_a\mid\mathcal D_a(u)
\sim\operatorname{Beta}\bigl(s_a+1,n_a-s_a+1\bigr),
\qquad
\mu_a(u)=\frac{s_a+1}{n_a+2}.
\label{eq:success-posterior}
\end{equation}

Let \(q(u)=\Pr(p_T>p_S\mid\mathcal D_T(u),\mathcal D_S(u))\) denote posterior support for a higher teacher completion probability (Figure~\ref{fig:method}). We combine its directional support with the positive posterior mean gap:

\begin{equation}
C(u)=
\underbrace{[2q(u)-1]_+}_{\text{directional support}}\times
\underbrace{[\mu_T(u)-\mu_S(u)]_+}_{\text{advantage magnitude}},
\qquad [z]_+=\max(z,0).
\label{eq:competence}
\end{equation}

The first factor measures directional support, and the mean gap measures advantage magnitude. Their product reduces supervision when either is small.

For TCOD-F2B, completion is measured at the teacher's full horizon and the student's current curriculum horizon (Appendix~\ref{sec:implementation}).

\hypertarget{method-outcome-conditioned-selection}{%
\subsubsection{Trajectory-level outcome selection}\label{method-outcome-conditioned-selection}}

Outcome-conditioned selection uses task completion to determine eligibility for direct teacher supervision (Figure~\ref{fig:method}, second row):

\begin{equation}
F(u,k)=\mathbf 1[Y_{S,u,k}=0]=1-Y_{S,u,k}.
\label{eq:outcome-selection}
\end{equation}

Here, \(\mathbf 1[\cdot]\) is the indicator function. Successful trajectories receive zero direct teacher weight and remain in the reward-based update. Failed trajectories are eligible for supervision, with weights further determined by competence and temporal coverage. All trajectories contribute to estimating \(C(u)\).

\hypertarget{method-temporal-supervision-prior}{%
\subsubsection{Turn-level temporal allocation}\label{method-temporal-supervision-prior}}

The temporal supervision prior uses successful teacher lengths to allocate guidance across action turns:

\begin{equation}
H_T(u,t)=
\begin{cases}
\displaystyle
\frac{1}{s_T(u)}
\sum_{i=1}^{n_T(u)}
Y_{T,u,i}\,\mathbf 1[L_{T,u,i}>t],
& s_T(u)>0,\\[6pt]
0,&s_T(u)=0.
\end{cases}
\label{eq:temporal-prior}
\end{equation}

For tasks with successful teacher rollouts, the temporal prior is the empirical fraction of successful teacher trajectories that contain action turn \(t\). It measures task-conditioned coverage over execution length and is applied within the base student rollout. Each factor lies in \([0,1]\), so their product also lies in \([0,1]\). ECOD composes this weight with the base method's configured teacher weighting and schedule.

\hypertarget{method-policy-optimization-with-ecod}{%
\subsection{Policy optimization with ECOD}\label{method-policy-optimization-with-ecod}}

ECOD uses the same allocation weight in teacher-loss and teacher-advantage formulations. Each formulation retains its policy-update computation and reward-advantage construction. We denote stop-gradient by \(\operatorname{sg}(\cdot)\).

\textbf{Teacher-loss weighting.} For OPD and TCOD-F2B, ECOD weights the teacher loss \(\ell^T_{u,k,t,j}(\theta)\) for response token \(j\) before aggregation:

\begin{equation}
\mathcal L_{\mathrm{ECOD}}(\theta)
=
\mathcal L_{\mathrm{RL}}(\theta)
+
\frac{\lambda}{N}
\sum_{(u,k,t,j)\in\mathcal I}
\operatorname{sg}\!\left[g(u,k,t)\right]\,
\ell^T_{u,k,t,j}(\theta),
\label{eq:teacher-loss}
\end{equation}

Here, \(\mathcal L_{\mathrm{RL}}\) includes the reward objective and existing regularizers, and \(\lambda\geq0\) is the base teacher coefficient. The set \(\mathcal I\) contains all valid response-token positions in a microbatch, with \(N=|\mathcal I|\), including zero-weight positions. Averaging over these tokens lets ECOD control both the distribution and the overall strength of supervision.

\textbf{Teacher-advantage weighting.} In our shared GRPO implementation, SOD and ATOD incorporate teacher feedback through a teacher-derived advantage. ECOD weights this contribution before combining it with the reward-derived advantage:

\begin{equation}
\widehat A^{\mathrm{ECOD}}_{u,k,t,j}
=
\widehat A^{\mathrm{RL}}_{u,k,t,j}
+
\operatorname{sg}\!\left[g(u,k,t)\right]\,
\widehat A^T_{u,k,t,j},
\label{eq:teacher-advantage}
\end{equation}

The reward and teacher terms use the coefficients configured for each experimental setting. The base policy objective is evaluated with the combined advantage. Appendix~\ref{sec:implementation} gives the implementation details, including the SOD adaptation; Algorithm~\ref{alg:ecod} in Appendix~\ref{sec:algorithm} gives the training procedure.

%% file: sections/experiments.tex
We evaluate ECOD through comparisons across RL and OPD methods, controlled allocation ablations, and training dynamics. These analyses assess integration across teacher objectives, selective allocation under matched supervision budgets, and teacher weighting during learning.

\hypertarget{experiments-experimental-setup}{%
\subsection{Experimental setup}\label{experiments-experimental-setup}}

\textbf{Benchmarks.} We use ALFWorld for household task completion~\citep{alfworld} and WebShop for instruction-guided shopping~\citep{webshop}. WebShop uses the 1,000-product configuration with synthetic goals. Appendix~\ref{appendix-benchmark-configurations} specifies the task splits, reset seeds, and interaction limits.

\textbf{Baselines.} We compare GRPO and GiGPO with their ECOD-assisted counterparts~\citep{grpo,gigpo}. We evaluate standard OPD on both backbones and use shared-framework implementations of SOD, TCOD-F2B, and ATOD in the multi-turn GRPO framework~\citep{sod,tcod,atod}. Each teacher-assisted method is evaluated with and without ECOD. Appendix~\ref{appendix-baseline-implementations} specifies the baseline implementations.

\textbf{Students and teachers.} Students are initialized from Qwen2.5-1.5B-Instruct or Qwen2.5-7B-Instruct and train for 150 updates~\citep{qwen25}. Paired teacher-assisted configurations share a frozen teacher of the same model size, trained for 150 updates with the corresponding benchmark and RL backbone. On 7B WebShop, ECOD-assisted runs use the teacher strengths specified in Appendix~\ref{appendix-training-details}; paired baselines retain their original strengths.

\textbf{Evaluation.} Validation is performed every five updates on 128 tasks. Main results use the designated checkpoint for each configuration, and auxiliary metrics use the same checkpoint. Ablations use update 150. Each configuration is trained once. Except for the seen entries detailed in Appendix~\ref{sec:evaluation}, reported means and sample standard deviations use eight decoding runs at fixed model weights. All evaluations use the student alone.

\hypertarget{experiments-main-results}{%
\subsection{Main results}\label{experiments-main-results}}

\label{sec:main-results}

\input{tables/main_results.tex}

Tables~\ref{tab:main-15b} and~\ref{tab:main-7b} compare the reported test results of each method and its ECOD-assisted counterpart.

\textbf{Allocation beyond standard distillation.} With the 1.5B student on ALFWorld unseen, GRPO-based OPD records 71.27\% success and its ECOD counterpart records 74.16\%. GiGPO-based OPD records 87.31\% and its ECOD counterpart records 89.55\%. These correspond to relative gains of 4.06\% and 2.56\%. Each pair shares the teacher and base teacher coefficient. Section~\ref{sec:ablation} separately evaluates allocation at a fixed training endpoint.

\textbf{Integration across distillation formulations.} The ECOD-assisted SOD, TCOD-F2B, and ATOD configurations have higher reported success in all six WebShop comparisons, with relative gains ranging from 2.29\% to 21.60\%. Across the two model scales, the mean paired relative changes are 6.20\% on ALFWorld unseen and 7.05\% on WebShop. These comparisons cover both teacher-loss and teacher-advantage implementations. Tables~\ref{tab:main-15b} and~\ref{tab:main-7b} report every paired result.

\textbf{Comparison with reward-based training.} With the 1.5B student, ECOD-assisted GRPO and GiGPO have higher reported success than their RL-only counterparts on both benchmarks. With the 7B student, GiGPO with ECOD records 91.32\% on ALFWorld unseen and 82.71\% on WebShop. These comparisons evaluate teacher-assisted training alongside GiGPO's step-level credit assignment.

\hypertarget{experiments-ablation-studies}{%
\subsection{Ablation studies}\label{experiments-ablation-studies}}

\label{sec:ablation}

We evaluate allocation factors and teacher-strength controls with Qwen2.5-1.5B-Instruct and GRPO at update 150 (Figure~\ref{fig:ablation}). Each WebShop control matches the weighted-token total that full ECOD would assign to the control's current rollout batch. It redistributes this total using an allocation rule with one factor removed, or with a pooled temporal prior.

\begingroup\setlength{\intextsep}{5pt}
\begin{figure}[!htbp]
\centering
\setlength{\abovecaptionskip}{5pt}
\includegraphics[width=\linewidth]{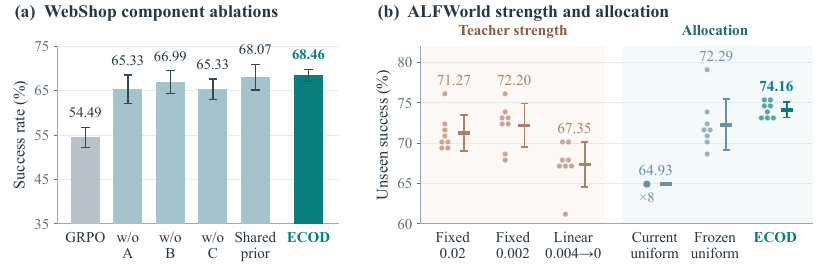}
\caption{\small \textbf{Allocation ablations.} (a) w/o A/B/C removes competence weighting/outcome selection/temporal weighting from the allocation pattern, retaining the ECOD-derived weighted-token total on each control's batch. (b) ALFWorld strength and uniform-allocation controls (Current/Frozen: current-batch/frozen budgets). Mean ± sample s.d. over eight decoding runs at step 150; dots show individual runs.}
\label{fig:ablation}
\end{figure}
\endgroup

\textbf{Allocation factors.} Full ECOD reaches 68.46\% mean WebShop success. Removing competence weighting, outcome selection, or temporal weighting from the allocation pattern gives 65.33\%, 66.99\%, and 65.33\%, respectively. A pooled temporal prior gives 68.07\%. Appendix~\ref{sec:app-ablation} defines the matching rule.

\textbf{Teacher strength and uniform allocation.} On ALFWorld unseen, full ECOD reaches 74.16\% mean success. Uniform OPD coefficients of 0.02 and 0.002 give 71.27\% and 72.20\%, while linear annealing gives 67.35\%. Current-batch and frozen-budget uniform allocation give 64.93\% and 72.29\%, respectively. Full ECOD has the highest mean success among these fixed-endpoint controls. Appendix~\ref{sec:teacher-strength-details} specifies the coefficient schedules and budget-matching procedures.

\hypertarget{experiments-teacher-supervision-and-learning-dynamics}{%
\subsection{Teacher weighting during training}\label{experiments-teacher-supervision-and-learning-dynamics}}

\label{sec:dynamics}

To examine learning dynamics, we compare teacher coefficients and validation success for SOD, TCOD-F2B, and ATOD on WebShop (Figure~\ref{fig:dynamics}).

\begingroup\setlength{\intextsep}{6pt}
\begin{figure}[!htbp]
\centering
\setlength{\abovecaptionskip}{5pt}
\includegraphics[width=\linewidth]{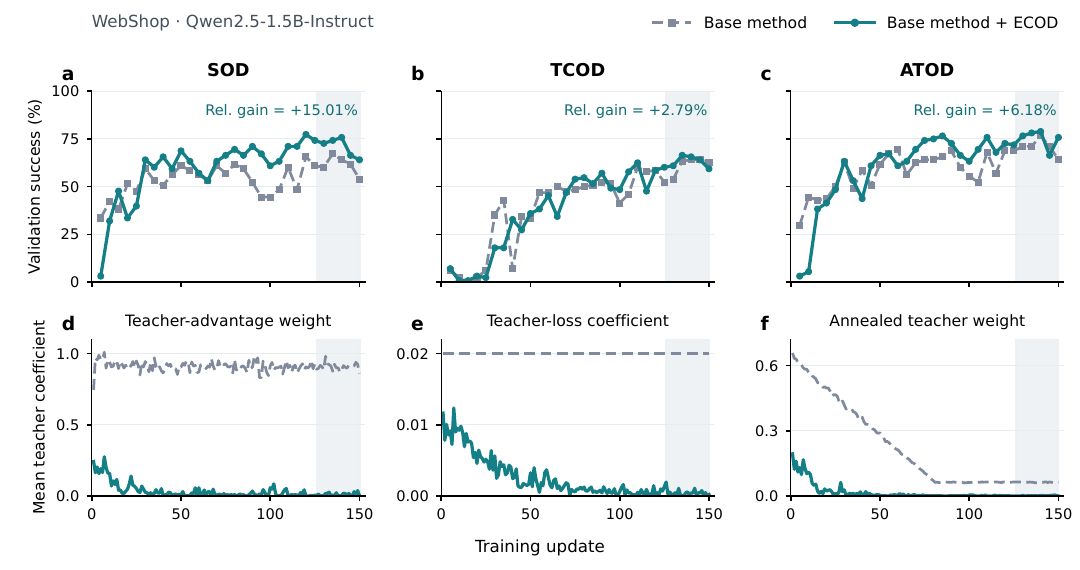}
\caption{\small \textbf{Learning dynamics on WebShop (1.5B, GRPO).} Dashed/solid curves denote base/ECOD methods: (a--c) validation success; (d--f) mean teacher coefficients on method-specific scales. Shading marks updates 126--150; annotations report relative gains in mean success. Evaluation follows Section~\ref{experiments-experimental-setup}; curve statistics are detailed in Appendix~\ref{sec:dynamics-protocol}.}
\label{fig:dynamics}
\end{figure}
\endgroup

Over updates 126--150, mean teacher coefficients under ECOD are 1.09\%, 2.08\%, and 1.38\% of the paired SOD, TCOD-F2B, and ATOD baselines. Mean success over the final five validation events is higher by 15.01\%, 2.79\%, and 6.18\%, respectively. The curves show higher late-stage success alongside lower teacher weighting. Appendix~\ref{sec:dynamics-protocol} defines the averaging protocol. Appendix~\ref{sec:gate-diagnostics} reports active-turn coverage and nonzero-weight distributions for the GRPO-based runs.

%% file: tables/main_results.tex
\definecolor{ECODGreen}{HTML}{0B5D3B}
\definecolor{ECODRed}{HTML}{AF443D}
\definecolor{ECODGray}{HTML}{66717B}
\begin{table}[!t]
\centering
\begingroup
\caption{Main results with Qwen2.5-1.5B-Instruct.}
\label{tab:main-15b}
\setlength{\tabcolsep}{2pt}
\renewcommand{\arraystretch}{1.05}
\fontsize{7.5}{9.5}\selectfont
\begin{tabular*}{\textwidth}{@{\extracolsep{\fill}}llrrrr@{}}
\toprule
& & \multicolumn{2}{c}{ALFWorld} & \multicolumn{2}{c}{WebShop} \\
\cmidrule(lr){3-4}\cmidrule(lr){5-6}
Baseline & Configuration & Seen SR (\%) $\uparrow$ & Unseen SR (\%) $\uparrow$ & SR (\%) $\uparrow$ & Task score (\%) $\uparrow$ \\
\midrule
\multirow{2}{*}{\textbf{GRPO}} & Base & $61.43$\,\makebox[29pt][r]{} & $57.56${\fontsize{6.5}{8}\selectfont$\pm 1.23$}\,\makebox[29pt][r]{} & $54.49${\fontsize{6.5}{8}\selectfont$\pm 2.28$}\,\makebox[29pt][r]{} & $64.46${\fontsize{6.5}{8}\selectfont$\pm 1.87$}\,\makebox[29pt][r]{} \\
 & \textbf{+ ECOD} & $75.00$\,\makebox[29pt][r]{\textcolor{ECODGreen}{$\uparrow\mathbf{22.09}$}} & $74.16${\fontsize{6.5}{8}\selectfont$\pm 0.97$}\,\makebox[29pt][r]{\textcolor{ECODGreen}{$\uparrow\mathbf{28.85}$}} & $68.46${\fontsize{6.5}{8}\selectfont$\pm 1.38$}\,\makebox[29pt][r]{\textcolor{ECODGreen}{$\uparrow\mathbf{25.63}$}} & $80.75${\fontsize{6.5}{8}\selectfont$\pm 1.71$}\,\makebox[29pt][r]{\textcolor{ECODGreen}{$\uparrow\mathbf{25.27}$}} \\
\midrule
\multirow{2}{*}{\textbf{GiGPO}} & Base & $82.86$\,\makebox[29pt][r]{} & $83.58${\fontsize{6.5}{8}\selectfont$\pm 0.00$}\,\makebox[29pt][r]{} & $73.24${\fontsize{6.5}{8}\selectfont$\pm 1.99$}\,\makebox[29pt][r]{} & $86.74${\fontsize{6.5}{8}\selectfont$\pm 1.57$}\,\makebox[29pt][r]{} \\
 & \textbf{+ ECOD} & $\mathbf{92.86}$\,\makebox[29pt][r]{\textcolor{ECODGreen}{$\uparrow\mathbf{12.07}$}} & $\mathbf{89.55}${\fontsize{6.5}{8}\selectfont$\pm 0.00$}\,\makebox[29pt][r]{\textcolor{ECODGreen}{$\uparrow\mathbf{7.14}$}} & $76.07${\fontsize{6.5}{8}\selectfont$\pm 1.67$}\,\makebox[29pt][r]{\textcolor{ECODGreen}{$\uparrow\mathbf{3.87}$}} & $\mathbf{89.42}${\fontsize{6.5}{8}\selectfont$\pm 0.85$}\,\makebox[29pt][r]{\textcolor{ECODGreen}{$\uparrow\mathbf{3.09}$}} \\
\midrule
\multirow{4}{*}{\textbf{OPD}} & GRPO & $75.00$\,\makebox[29pt][r]{} & $71.27${\fontsize{6.5}{8}\selectfont$\pm 2.22$}\,\makebox[29pt][r]{} & $68.26${\fontsize{6.5}{8}\selectfont$\pm 2.04$}\,\makebox[29pt][r]{} & $77.39${\fontsize{6.5}{8}\selectfont$\pm 1.95$}\,\makebox[29pt][r]{} \\
 & \textbf{GRPO + ECOD} & $75.00$\,\makebox[29pt][r]{\textcolor{ECODGray}{$\leftrightarrow\mathbf{0.00}$}} & $74.16${\fontsize{6.5}{8}\selectfont$\pm 0.97$}\,\makebox[29pt][r]{\textcolor{ECODGreen}{$\uparrow\mathbf{4.06}$}} & $68.46${\fontsize{6.5}{8}\selectfont$\pm 1.38$}\,\makebox[29pt][r]{\textcolor{ECODGreen}{$\uparrow\mathbf{0.29}$}} & $80.75${\fontsize{6.5}{8}\selectfont$\pm 1.71$}\,\makebox[29pt][r]{\textcolor{ECODGreen}{$\uparrow\mathbf{4.35}$}} \\
 & GiGPO & $\mathbf{92.86}$\,\makebox[29pt][r]{} & $87.31${\fontsize{6.5}{8}\selectfont$\pm 0.00$}\,\makebox[29pt][r]{} & $76.07${\fontsize{6.5}{8}\selectfont$\pm 1.32$}\,\makebox[29pt][r]{} & $89.31${\fontsize{6.5}{8}\selectfont$\pm 1.12$}\,\makebox[29pt][r]{} \\
 & \textbf{GiGPO + ECOD} & $\mathbf{92.86}$\,\makebox[29pt][r]{\textcolor{ECODGray}{$\leftrightarrow\mathbf{0.00}$}} & $\mathbf{89.55}${\fontsize{6.5}{8}\selectfont$\pm 0.00$}\,\makebox[29pt][r]{\textcolor{ECODGreen}{$\uparrow\mathbf{2.56}$}} & $76.07${\fontsize{6.5}{8}\selectfont$\pm 1.67$}\,\makebox[29pt][r]{\textcolor{ECODGray}{$\leftrightarrow\mathbf{0.00}$}} & $\mathbf{89.42}${\fontsize{6.5}{8}\selectfont$\pm 0.85$}\,\makebox[29pt][r]{\textcolor{ECODGreen}{$\uparrow\mathbf{0.12}$}} \\
\midrule
\multirow{2}{*}{\textbf{SOD}} & Base & $65.00$\,\makebox[29pt][r]{} & $66.98${\fontsize{6.5}{8}\selectfont$\pm 1.25$}\,\makebox[29pt][r]{} & $62.40${\fontsize{6.5}{8}\selectfont$\pm 2.02$}\,\makebox[29pt][r]{} & $74.27${\fontsize{6.5}{8}\selectfont$\pm 2.81$}\,\makebox[29pt][r]{} \\
 & \textbf{+ ECOD} & $75.71$\,\makebox[29pt][r]{\textcolor{ECODGreen}{$\uparrow\mathbf{16.48}$}} & $81.34${\fontsize{6.5}{8}\selectfont$\pm 0.00$}\,\makebox[29pt][r]{\textcolor{ECODGreen}{$\uparrow\mathbf{21.45}$}} & $75.88${\fontsize{6.5}{8}\selectfont$\pm 1.06$}\,\makebox[29pt][r]{\textcolor{ECODGreen}{$\uparrow\mathbf{21.60}$}} & $88.23${\fontsize{6.5}{8}\selectfont$\pm 0.87$}\,\makebox[29pt][r]{\textcolor{ECODGreen}{$\uparrow\mathbf{18.80}$}} \\
\midrule
\multirow{2}{*}{\textbf{TCOD}} & F2B & $70.00$\,\makebox[29pt][r]{} & $70.34${\fontsize{6.5}{8}\selectfont$\pm 1.25$}\,\makebox[29pt][r]{} & $68.16${\fontsize{6.5}{8}\selectfont$\pm 3.06$}\,\makebox[29pt][r]{} & $79.06${\fontsize{6.5}{8}\selectfont$\pm 2.63$}\,\makebox[29pt][r]{} \\
 & \textbf{F2B + ECOD} & $80.00$\,\makebox[29pt][r]{\textcolor{ECODGreen}{$\uparrow\mathbf{14.29}$}} & $79.10${\fontsize{6.5}{8}\selectfont$\pm 0.98$}\,\makebox[29pt][r]{\textcolor{ECODGreen}{$\uparrow\mathbf{12.47}$}} & $69.73${\fontsize{6.5}{8}\selectfont$\pm 2.35$}\,\makebox[29pt][r]{\textcolor{ECODGreen}{$\uparrow\mathbf{2.29}$}} & $82.03${\fontsize{6.5}{8}\selectfont$\pm 1.86$}\,\makebox[29pt][r]{\textcolor{ECODGreen}{$\uparrow\mathbf{3.76}$}} \\
\midrule
\multirow{2}{*}{\textbf{ATOD}} & Base & $74.29$\,\makebox[29pt][r]{} & $78.64${\fontsize{6.5}{8}\selectfont$\pm 1.05$}\,\makebox[29pt][r]{} & $71.88${\fontsize{6.5}{8}\selectfont$\pm 1.32$}\,\makebox[29pt][r]{} & $84.59${\fontsize{6.5}{8}\selectfont$\pm 1.06$}\,\makebox[29pt][r]{} \\
 & \textbf{+ ECOD} & $80.71$\,\makebox[29pt][r]{\textcolor{ECODGreen}{$\uparrow\mathbf{8.65}$}} & $73.51${\fontsize{6.5}{8}\selectfont$\pm 1.06$}\,\makebox[29pt][r]{\textcolor{ECODRed}{$\downarrow\mathbf{6.52}$}} & $\mathbf{77.83}${\fontsize{6.5}{8}\selectfont$\pm 1.38$}\,\makebox[29pt][r]{\textcolor{ECODGreen}{$\uparrow\mathbf{8.29}$}} & $87.89${\fontsize{6.5}{8}\selectfont$\pm 0.88$}\,\makebox[29pt][r]{\textcolor{ECODGreen}{$\uparrow\mathbf{3.89}$}} \\
\bottomrule
\end{tabular*}\par
\endgroup
\begingroup
\caption{Main results with Qwen2.5-7B-Instruct.}
\label{tab:main-7b}
\setlength{\tabcolsep}{2pt}
\renewcommand{\arraystretch}{1.05}
\fontsize{7.5}{9.5}\selectfont
\begin{tabular*}{\textwidth}{@{\extracolsep{\fill}}llrrrr@{}}
\toprule
& & \multicolumn{2}{c}{ALFWorld} & \multicolumn{2}{c}{WebShop} \\
\cmidrule(lr){3-4}\cmidrule(lr){5-6}
Baseline & Configuration & Seen SR (\%) $\uparrow$ & Unseen SR (\%) $\uparrow$ & SR (\%) $\uparrow$ & Task score (\%) $\uparrow$ \\
\midrule
\multirow{2}{*}{\textbf{GRPO}} & Base & $77.14$\,\makebox[29pt][r]{} & $72.39${\fontsize{6.5}{8}\selectfont$\pm 0.00$}\,\makebox[29pt][r]{} & $73.34${\fontsize{6.5}{8}\selectfont$\pm 3.67$}\,\makebox[29pt][r]{} & $83.10${\fontsize{6.5}{8}\selectfont$\pm 2.64$}\,\makebox[29pt][r]{} \\
 & \textbf{+ ECOD} & $77.14$\,\makebox[29pt][r]{\textcolor{ECODGray}{$\leftrightarrow\mathbf{0.00}$}} & $72.39${\fontsize{6.5}{8}\selectfont$\pm 0.00$}\,\makebox[29pt][r]{\textcolor{ECODGray}{$\leftrightarrow\mathbf{0.00}$}} & $75.39${\fontsize{6.5}{8}\selectfont$\pm 1.32$}\,\makebox[29pt][r]{\textcolor{ECODGreen}{$\uparrow\mathbf{2.80}$}} & $86.05${\fontsize{6.5}{8}\selectfont$\pm 1.04$}\,\makebox[29pt][r]{\textcolor{ECODGreen}{$\uparrow\mathbf{3.55}$}} \\
\midrule
\multirow{2}{*}{\textbf{GiGPO}} & Base & $89.29$\,\makebox[29pt][r]{} & $87.50${\fontsize{6.5}{8}\selectfont$\pm 0.66$}\,\makebox[29pt][r]{} & $74.51${\fontsize{6.5}{8}\selectfont$\pm 1.38$}\,\makebox[29pt][r]{} & $84.56${\fontsize{6.5}{8}\selectfont$\pm 1.96$}\,\makebox[29pt][r]{} \\
 & \textbf{+ ECOD} & $\mathbf{95.71}$\,\makebox[29pt][r]{\textcolor{ECODGreen}{$\uparrow\mathbf{7.20}$}} & $\mathbf{91.32}${\fontsize{6.5}{8}\selectfont$\pm 0.79$}\,\makebox[29pt][r]{\textcolor{ECODGreen}{$\uparrow\mathbf{4.37}$}} & $\mathbf{82.71}${\fontsize{6.5}{8}\selectfont$\pm 0.88$}\,\makebox[29pt][r]{\textcolor{ECODGreen}{$\uparrow\mathbf{11.01}$}} & $\mathbf{92.97}${\fontsize{6.5}{8}\selectfont$\pm 0.72$}\,\makebox[29pt][r]{\textcolor{ECODGreen}{$\uparrow\mathbf{9.94}$}} \\
\midrule
\multirow{4}{*}{\textbf{OPD}} & GRPO & $75.71$\,\makebox[29pt][r]{} & $72.39${\fontsize{6.5}{8}\selectfont$\pm 2.15$}\,\makebox[29pt][r]{} & $77.05${\fontsize{6.5}{8}\selectfont$\pm 1.32$}\,\makebox[29pt][r]{} & $85.91${\fontsize{6.5}{8}\selectfont$\pm 1.11$}\,\makebox[29pt][r]{} \\
 & \textbf{GRPO + ECOD} & $77.14$\,\makebox[29pt][r]{\textcolor{ECODGreen}{$\uparrow\mathbf{1.89}$}} & $72.39${\fontsize{6.5}{8}\selectfont$\pm 0.00$}\,\makebox[29pt][r]{\textcolor{ECODGray}{$\leftrightarrow\mathbf{0.00}$}} & $75.39${\fontsize{6.5}{8}\selectfont$\pm 1.32$}\,\makebox[29pt][r]{\textcolor{ECODRed}{$\downarrow\mathbf{2.15}$}} & $86.05${\fontsize{6.5}{8}\selectfont$\pm 1.04$}\,\makebox[29pt][r]{\textcolor{ECODGreen}{$\uparrow\mathbf{0.17}$}} \\
 & GiGPO & $93.57$\,\makebox[29pt][r]{} & $90.11${\fontsize{6.5}{8}\selectfont$\pm 1.31$}\,\makebox[29pt][r]{} & $80.47${\fontsize{6.5}{8}\selectfont$\pm 1.02$}\,\makebox[29pt][r]{} & $88.99${\fontsize{6.5}{8}\selectfont$\pm 0.90$}\,\makebox[29pt][r]{} \\
 & \textbf{GiGPO + ECOD} & $\mathbf{95.71}$\,\makebox[29pt][r]{\textcolor{ECODGreen}{$\uparrow\mathbf{2.29}$}} & $\mathbf{91.32}${\fontsize{6.5}{8}\selectfont$\pm 0.79$}\,\makebox[29pt][r]{\textcolor{ECODGreen}{$\uparrow\mathbf{1.35}$}} & $\mathbf{82.71}${\fontsize{6.5}{8}\selectfont$\pm 0.88$}\,\makebox[29pt][r]{\textcolor{ECODGreen}{$\uparrow\mathbf{2.79}$}} & $\mathbf{92.97}${\fontsize{6.5}{8}\selectfont$\pm 0.72$}\,\makebox[29pt][r]{\textcolor{ECODGreen}{$\uparrow\mathbf{4.46}$}} \\
\midrule
\multirow{2}{*}{\textbf{SOD}} & Base & $79.29$\,\makebox[29pt][r]{} & $75.56${\fontsize{6.5}{8}\selectfont$\pm 1.04$}\,\makebox[29pt][r]{} & $79.88${\fontsize{6.5}{8}\selectfont$\pm 2.04$}\,\makebox[29pt][r]{} & $87.50${\fontsize{6.5}{8}\selectfont$\pm 1.14$}\,\makebox[29pt][r]{} \\
 & \textbf{+ ECOD} & $82.86$\,\makebox[29pt][r]{\textcolor{ECODGreen}{$\uparrow\mathbf{4.50}$}} & $75.37${\fontsize{6.5}{8}\selectfont$\pm 0.00$}\,\makebox[29pt][r]{\textcolor{ECODRed}{$\downarrow\mathbf{0.25}$}} & $82.03${\fontsize{6.5}{8}\selectfont$\pm 1.82$}\,\makebox[29pt][r]{\textcolor{ECODGreen}{$\uparrow\mathbf{2.69}$}} & $90.28${\fontsize{6.5}{8}\selectfont$\pm 1.26$}\,\makebox[29pt][r]{\textcolor{ECODGreen}{$\uparrow\mathbf{3.18}$}} \\
\midrule
\multirow{2}{*}{\textbf{TCOD}} & F2B & $81.43$\,\makebox[29pt][r]{} & $74.63${\fontsize{6.5}{8}\selectfont$\pm 0.00$}\,\makebox[29pt][r]{} & $77.54${\fontsize{6.5}{8}\selectfont$\pm 1.30$}\,\makebox[29pt][r]{} & $86.76${\fontsize{6.5}{8}\selectfont$\pm 1.18$}\,\makebox[29pt][r]{} \\
 & \textbf{F2B + ECOD} & $84.29$\,\makebox[29pt][r]{\textcolor{ECODGreen}{$\uparrow\mathbf{3.51}$}} & $81.34${\fontsize{6.5}{8}\selectfont$\pm 1.60$}\,\makebox[29pt][r]{\textcolor{ECODGreen}{$\uparrow\mathbf{9.00}$}} & $79.88${\fontsize{6.5}{8}\selectfont$\pm 0.91$}\,\makebox[29pt][r]{\textcolor{ECODGreen}{$\uparrow\mathbf{3.02}$}} & $88.47${\fontsize{6.5}{8}\selectfont$\pm 1.01$}\,\makebox[29pt][r]{\textcolor{ECODGreen}{$\uparrow\mathbf{1.98}$}} \\
\midrule
\multirow{2}{*}{\textbf{ATOD}} & Base & $83.57$\,\makebox[29pt][r]{} & $79.10${\fontsize{6.5}{8}\selectfont$\pm 0.00$}\,\makebox[29pt][r]{} & $79.20${\fontsize{6.5}{8}\selectfont$\pm 1.72$}\,\makebox[29pt][r]{} & $87.15${\fontsize{6.5}{8}\selectfont$\pm 1.40$}\,\makebox[29pt][r]{} \\
 & \textbf{+ ECOD} & $85.71$\,\makebox[29pt][r]{\textcolor{ECODGreen}{$\uparrow\mathbf{2.56}$}} & $79.94${\fontsize{6.5}{8}\selectfont$\pm 1.93$}\,\makebox[29pt][r]{\textcolor{ECODGreen}{$\uparrow\mathbf{1.06}$}} & $\mathbf{82.71}${\fontsize{6.5}{8}\selectfont$\pm 0.77$}\,\makebox[29pt][r]{\textcolor{ECODGreen}{$\uparrow\mathbf{4.44}$}} & $89.84${\fontsize{6.5}{8}\selectfont$\pm 0.87$}\,\makebox[29pt][r]{\textcolor{ECODGreen}{$\uparrow\mathbf{3.08}$}} \\
\bottomrule
\end{tabular*}\par
\endgroup

\begin{minipage}{\linewidth}\footnotesize SR: success rate; task score: normalized reward, both expressed as percentages. Each row reports the designated checkpoint for that configuration. Unseen and WebShop metrics report means and sample standard deviations over eight decoding runs at fixed weights. Seen entries omit standard deviations; their repeat counts are specified in Appendix A.8. Arrows give relative changes from paired baselines using unrounded means. Bold marks the best entry in each column. The 7B WebShop ECOD configurations use the adjusted teacher strengths in Section 5.1. SOD denotes the shared-framework implementation in Appendix A.6.\end{minipage}
\end{table}

%% file: sections/conclusion.tex
ECOD allocates teacher supervision from task-matched completion outcomes and successful teacher execution lengths. It combines task-level competence calibration, trajectory-level outcome selection, and turn-level temporal allocation in a weight applied to teacher losses or teacher-derived advantages. Experiments on ALFWorld and WebShop cover two model scales and multiple RL and OPD implementations. Fixed-endpoint controls show higher mean completion success for full ECOD than for the tested coefficient schedules and uniform-allocation rules. The trained student acts without teacher intervention at inference.

%% file: sections/ai_use_statement.tex
In this work, we used generative AI tools to assist with literature search and background research, as well as language editing and stylistic refinement of author-written manuscript drafts. We also used generative AI tools to assist with portions of coding and the implementation and refinement of the experimental pipeline. The research ideas, methodological design, experimental procedures, selection of benchmarks, baselines, datasets, and models, as well as the initial experimental framework, were developed by the authors. AI tools were used primarily to help translate these designs into code and improve their implementation. All AI-assisted text, code, and experimental components were reviewed, revised, and verified by the authors. We take full responsibility for the final content and results of this work.

%% file: sections/appendix.tex
\section{Supplementary material}\label{appendix-experimental-details}
\label{sec:appendix}

\setcounter{table}{0}
\renewcommand{\thetable}{A\arabic{table}}
\renewcommand{\theHtable}{A\arabic{table}}
\setcounter{figure}{0}
\renewcommand{\thefigure}{A\arabic{figure}}
\renewcommand{\theHfigure}{A\arabic{figure}}

\noindent\textbf{Appendix contents}\par
\begingroup
\small\setlength{\parskip}{2pt}
\newcommand{\appendixentry}[1]{\noindent\hyperref[#1]{\ref*{#1}\quad\nameref*{#1}}\dotfill\pageref{#1}\par}
\appendixentry{sec:motivation-protocol}
\appendixentry{sec:recent-opd-comparison}
\appendixentry{sec:implementation}
\appendixentry{sec:algorithm}
\appendixentry{appendix-benchmark-configurations}
\appendixentry{appendix-baseline-implementations}
\appendixentry{appendix-training-details}
\appendixentry{sec:evaluation}
\appendixentry{sec:app-ablation}
\appendixentry{sec:teacher-strength-details}
\appendixentry{sec:dynamics-protocol}
\appendixentry{sec:gate-diagnostics}
\endgroup
\input{sections/motivation_protocol}

\input{sections/recent_opd_comparison}

\hypertarget{appendix-ecod-implementation}{%
\subsection{ECOD implementation}\label{appendix-ecod-implementation}}

\label{sec:implementation}

Each turn weight \(g(u,k,t)\) applies to all valid reasoning and action tokens in that response, excluding environment observations and padding. Teacher and student predictions share the same interaction history and response prefix. Allocation weights are detached during policy optimization.

\hypertarget{appendix-rollout-statistics-and-posterior-comparison}{%
\subsubsection{Rollout statistics and posterior comparison}\label{appendix-rollout-statistics-and-posterior-comparison}}

Teacher statistics are collected once from the frozen teacher. Student statistics are recomputed from each new rollout batch under the rollout policy \(\pi_{\mathrm{old}}\). Each trajectory contributes one binary completion outcome. Completion is assessed at each policy's collection horizon; for TCOD-F2B, the teacher uses the full horizon and the student uses its current curriculum horizon (Appendix~\ref{appendix-training-details}).

Let \(a_T=s_T+1\), \(b_T=n_T-s_T+1\), \(a_S=s_S+1\), and \(b_S=n_S-s_S+1\). The integer Beta shapes permit deterministic evaluation of the posterior comparison:

\begin{equation}
q(u)=\sum_{i=0}^{a_T-1}
\frac{B(a_S+i,b_T+b_S)}{(b_T+i)B(1+i,b_T)B(a_S,b_S)}.
\label{eq:posterior-comparison}
\end{equation}

Here, \(B\) is the beta function. The implementation evaluates each summand through log-beta values computed with log-gamma functions, caches repeated count combinations, and clips numerical results to \([0,1]\). The temporal table stores the empirical survival fraction for every required task--turn pair, including genuine zeros. Missing entries are treated as configuration errors rather than zero temporal support.

\hypertarget{appendix-explicit-teacher-loss}{%
\subsubsection{Explicit teacher loss}\label{appendix-explicit-teacher-loss}}

The explicit teacher loss uses the importance-corrected sampled surrogate. For each sampled response token, let \(z_\theta\), \(z_{\mathrm{old}}\), and \(z_T\) denote the current-student, rollout-student, and teacher log probabilities on the same context. Teacher and rollout-student scores are detached. The implemented surrogate is

\begin{equation}
\begin{aligned}
d&=\operatorname{clip}(z_\theta-z_T,-20,20),
&r&=\exp\!\left(\operatorname{clip}(z_\theta-z_{\mathrm{old}},-20,20)\right),\\
v&=\operatorname{clip}\!\left(r(e^{-d}-1+d),0,10\right),
&\ell^T&=\operatorname{sg}(v)+\operatorname{sg}(rd)\left(z_\theta-\operatorname{sg}(z_\theta)\right).
\end{aligned}
\label{eq:teacher-surrogate}
\end{equation}

The second term has zero forward value and supplies the gradient \(\operatorname{sg}(rd)\) times the current-student log-probability gradient. Without clipping, this equals the gradient of the importance-weighted sampled expression with the sampled context held fixed. The forward cap applies to \(v\); the detached gradient coefficient uses the clipped log ratios. ECOD multiplies the surrogate by its detached turn weight before reduction.

\hypertarget{appendix-reduction-and-parameter-updates}{%
\subsubsection{Reduction and parameter updates}\label{appendix-reduction-and-parameter-updates}}

The explicit teacher term uses \texttt{all\_response\_token\_mean} within each device-local microbatch. Its denominator counts all valid response tokens, including zero-weight positions, and excludes environment observations and padding. A microbatch with zero total supervision contributes zero teacher loss. Training averages these microbatch objectives across gradient-accumulation steps and retains the base FSDP gradient reduction. Microbatch means receive equal accumulation weights, independently of their token counts. The same reduction is used by each paired explicit-loss baseline and its ECOD extension.

\hypertarget{appendix-teacher-derived-advantages}{%
\subsubsection{Teacher-derived advantages}\label{appendix-teacher-derived-advantages}}

For SOD and ATOD, the teacher token signal is \(\Delta_j=z_{T,j}-z_{\mathrm{old},j}\), computed once per rollout batch with detached scores. Let \(w_t\) denote the base method's turn weight. SOD obtains \(w_t\) from its divergence-ratio rule; ATOD computes it from within-trajectory normalized disagreement and uncertainty. ECOD acts after these native weights have been constructed:

\begin{equation}
\begin{aligned}
\widehat A^{\mathrm{SOD+ECOD}}_{t,j}
&=\widehat A^{\mathrm{RL}}_{t,j}+g_t\,\eta w_t\Delta_j,\\
\widehat A^{\mathrm{ATOD+ECOD}}_{t,j}
&=\rho_m\widehat A^{\mathrm{RL}}_{t,j}+g_t\,\kappa_m w_t\Delta_j.
\end{aligned}
\label{eq:variant-advantages}
\end{equation}

Task and trajectory indices are suppressed. Here, \(\eta\) is the configured SOD teacher strength, and \(\rho_m,\kappa_m\) follow the ATOD schedule at update \(m\). The 1.5B SOD implementation uses \(\eta=1\); the 7B WebShop adjustment is specified in Appendix~\ref{appendix-training-details}. Reward-derived advantages are computed and normalized by the base method before this combination. There is no additional normalization of the combined advantage before the actor's PPO objective. ECOD preserves the construction and normalization of reward-derived advantages and retains native schedules; PPO clipping is subsequently applied to the combined advantage. All advantages and ECOD weights remain fixed during optimization of the rollout batch.

\hypertarget{appendix-training-algorithm}{%
\subsection{Training algorithm}\label{appendix-training-algorithm}}

\label{sec:algorithm}

\begin{algorithm}[H]
\caption{Training with ECOD}\label{alg:ecod}
\small\begin{algorithmic}[1]
\Statex Inputs: student $\pi_\theta$, fixed teacher $\pi_T$, training task conditions, base policy optimizer with a separable teacher contribution
\Statex Prepare: collect teacher rollouts; cache success counts and successful lengths
\Statex For each training update:
\State Set $\pi_{\mathrm{old}}$ $\leftarrow$ $\pi_\theta$; collect student rollouts for the sampled task conditions.
\State Record one completion outcome per rollout; compute student counts.
\State Compare teacher/student posteriors to obtain $C(u)$.
\State Construct $F(u,k)$ from student outcomes and query $H_T(u,t)$.
\State Form $g(u,k,t)=C(u)\times F(u,k)\times H_T(u,t)$; detach the coefficients.
\State Evaluate teacher feedback on the student-generated token contexts.
\State Weight the teacher loss {[}Eq.~\eqref{eq:teacher-loss}{]} or teacher advantage {[}Eq.~\eqref{eq:teacher-advantage}{]} before aggregation or combination with the reward-derived term.
\State Update $\theta$ with the base optimizer, keeping $g$ fixed for this batch.
\Statex Output: student policy $\pi_\theta$; no teacher is required at inference.
\end{algorithmic}\end{algorithm}

Cached teacher rollouts provide allocation statistics; online teacher scoring supplies the token-level training signal.

\hypertarget{appendix-benchmark-configurations}{%
\subsection{Benchmark configurations}\label{appendix-benchmark-configurations}}

\textbf{ALFWorld.} We used ALFWorld's text-based TextWorld environment~\citep{alfworld}. Following the environment filtering in our verl-agent implementation, the task pools contained 3,553 training games, 140 seen evaluation games, and 134 unseen evaluation games. We included all six task types: Pick, Look, Clean, Heat, Cool, and Pick2. The interaction horizon was capped at 50 action turns, with the two most recent observation--action turns retained in the history.

\textbf{WebShop.} We used the 1,000-product, synthetic-goal WebShop configuration in verl-agent~\citep{webshop}. Training and evaluation reset seeds were sampled from the disjoint ranges \([0,2^{16})\) and \([2^{16},2^{32})\), respectively. Final evaluation used a fixed schedule of 128 task resets. We allowed at most 15 action turns and retained the two most recent observation--action turns.

\hypertarget{appendix-baseline-implementations}{%
\subsection{Baseline implementations}\label{appendix-baseline-implementations}}

We organized the baselines into two groups. GRPO and GiGPO provide RL-only reference points, whereas OPD, SOD, TCOD, and ATOD examine different uses of teacher supervision.

\begin{itemize}
\tightlist
\item
  \textbf{GRPO:} Group Relative Policy Optimization uses rewards from groups of sampled responses to construct relative advantages. We used its multi-turn implementation in verl-agent as a teacher-free baseline~\citep{grpo}.
\item
  \textbf{GiGPO:} Group-in-Group Policy Optimization combines episode-level comparison with step-level grouping of actions from shared environment states, providing a second RL backbone with finer-grained credit assignment~\citep{gigpo}.
\item
  \textbf{OPD:} On-policy distillation evaluates the teacher on student-generated contexts. Our OPD baseline added a teacher-KL term to the underlying RL objective and applied it to all valid response tokens, including those from successful trajectories. This served as the direct comparison for ECOD's supervision allocation~\citep{gkd}.
\item
  \textbf{SOD:} Step-wise On-policy Distillation adjusts the strength of distillation according to teacher--student divergence at each interaction step. We use our shared-framework implementation of its step-weighted teacher signal within the shared multi-turn GRPO training system~\citep{sod}.
\item
  \textbf{TCOD:} Temporal Curriculum On-Policy Distillation progressively expands the depth of student interaction. We implemented its front-to-back curriculum, denoted \textbf{TCOD-F2B}, by increasing the training rollout horizon while retaining the GRPO-plus-OPD objective~\citep{tcod}.
\item
  \textbf{ATOD:} Annealed Turn-Aware On-Policy Distillation combines an annealed balance between OPD and RL with turn-level disagreement--uncertainty reweighting. We implemented these teacher-side mechanisms on the same multi-turn GRPO backbone~\citep{atod}.
\end{itemize}

The core comparison evaluated RL alone, RL with OPD, and RL with ECOD under both GRPO and GiGPO. The compatibility comparisons evaluated the specified SOD, TCOD-F2B, and ATOD mechanisms with and without ECOD in a shared multi-turn GRPO framework. In each pair, ECOD acted only on the teacher contribution; the environment-derived reward signal and the base method's curriculum or weighting rule were retained.

The same ECOD runs are compared against the RL-only and OPD baselines in the two table groups (Tables~\ref{tab:main-15b} and~\ref{tab:main-7b}).

\hypertarget{appendix-training-details}{%
\subsection{Training details}\label{appendix-training-details}}

\textbf{Student models and hardware.} Students were initialized from Qwen2.5-1.5B-Instruct and Qwen2.5-7B-Instruct for experiments on ALFWorld and WebShop. Each run used a single node with eight NVIDIA H200 GPUs. Our implementation used verl-agent, built on verl \citep{hybridflow}, with FSDP for distributed policy optimization and vLLM~\citep{vllm} for rollout generation. The training schedule comprised 150 updates, each with 16 task conditions and eight rollouts per condition, yielding 128 trajectories per update. We used a learning rate of \(10^{-6}\), a training sampling temperature of 1.0, and one PPO epoch per update. Table A2 summarizes the Qwen2.5-1.5B-Instruct settings and benchmark-specific differences.

\textbf{Teacher construction.} Teachers were trained from Qwen2.5-1.5B-Instruct or Qwen2.5-7B-Instruct, matching the student's model size. For each benchmark, model size, and RL backbone, we froze the teacher obtained from the corresponding RL-only run after 150 updates. GRPO-based OPD, SOD, TCOD-F2B, ATOD, and their ECOD-assisted variants used the GRPO-trained teacher; GiGPO-based configurations used the GiGPO-trained teacher. Teacher-assisted students started from the common pretrained model, not from the teacher weights. The teacher generated eight rollouts for each entry in the frozen \(150\times16\) training task-condition schedule. ECOD used their completion outcomes and successful trajectory lengths, together with student outcomes from the current batch, to construct its allocation weights. Teacher-assisted variants shared the teacher within each benchmark, model size, and RL backbone; paired comparisons reused the training and validation task schedules. Teacher precollection and online teacher scoring incurred additional cost beyond the student rollout budget.

\textbf{Objectives and matching.} For Qwen2.5-1.5B-Instruct, the RL-only, OPD, ECOD, and TCOD-F2B configurations retained a reference-policy KL coefficient of 0.01; those using an explicit teacher-KL term used a separate coefficient of 0.02. SOD and ATOD instead incorporated teacher feedback into their advantages, without an additional reference-KL or separate teacher-KL loss. Within these pairs, the teacher and non-ECOD hyperparameters were kept the same. Update and rollout counts are matched, while action and token consumption vary with TCOD-F2B's curriculum horizon.

\textbf{7B WebShop teacher strengths.} Experiments with Qwen2.5-7B-Instruct on WebShop used the same task configuration, update budget, and rollout group size, with a teacher of the same size trained using the corresponding RL backbone. ECOD used a teacher-KL coefficient of 0.08 for GRPO, GiGPO, and TCOD-F2B, a teacher-advantage coefficient of 4.0 for SOD, and four times the ATOD \(\kappa\) schedule. The corresponding baselines retained their original teacher-signal strengths.

\textbf{Table A2. Training settings for the Qwen2.5-1.5B-Instruct experiments. Mini-batch sizes count response-turn samples, not complete trajectories.}

\begin{longtable}[]{@{}lll@{}}
\toprule
Setting & ALFWorld & WebShop \\
\midrule
\endhead
Student initialization & Qwen2.5-1.5B-Instruct & Qwen2.5-1.5B-Instruct \\
Hardware & 1 node × 8 NVIDIA H200 & 1 node × 8 NVIDIA H200 \\
Training updates & 150 & 150 \\
Task conditions per update & 16 & 16 \\
Rollouts per condition & 8 & 8 \\
Trajectories per update & 128 & 128 \\
Learning rate & \(10^{-6}\) & \(10^{-6}\) \\
PPO epochs per update & 1 & 1 \\
PPO mini-batch size & 256 & 64 \\
PPO clip ratio & 0.2 & 0.2 \\
Entropy coefficient & 0.001 & 0.001 \\
Maximum action turns & 50 & 15 \\
History length & 2 & 2 \\
Maximum prompt length per turn & 2,048 tokens & 4,096 tokens \\
Maximum response length per turn & 512 tokens & 512 tokens \\
Training / evaluation temperature & 1.0 / 0.4 & 1.0 / 0.4 \\
Validation frequency & Every 5 updates & Every 5 updates \\
Validation batch & 128 & 128 \\
Invalid-action penalty & 0.1 & 0.1 \\
GiGPO discount \(\gamma\) & 0.95 & 0.95 \\
GiGPO step-advantage weight & 1.0 & 1.0 \\
GiGPO normalization & mean\_std\_norm & mean\_norm \\
\bottomrule
\end{longtable}

The maximum action horizon in Table A2 applies throughout evaluation and to non-curriculum training. For TCOD-F2B training, the horizon at update \(m\geq1\) was \(K_m=\min(T_{\max},1+\lfloor m/2\rfloor)\), where \(T_{\max}\) is the benchmark limit. Teacher statistics were collected at the full horizon.

\textbf{Table A3. Method-specific settings for the 1.5B comparisons. The 7B WebShop teacher-strength adjustments are specified above.}

\begin{longtable}[]{@{}
  >{\raggedright\arraybackslash}p{(\columnwidth - 2\tabcolsep) * \real{0.50}}
  >{\raggedright\arraybackslash}p{(\columnwidth - 2\tabcolsep) * \real{0.50}}@{}}
\toprule
Component & Setting \\
\midrule
\endhead
Explicit teacher-KL objectives & Reference KL coefficient 0.01; teacher KL coefficient 0.02 \\
SOD & Divergence-ratio stabilization \(\epsilon=10^{-6}\); weight cap \(1+\delta\), \(\delta=0.2\) \\
TCOD-F2B & Initial horizon 1; growth interval 2 updates; benchmark-specific cap \\
ATOD & OPD weight \(\kappa:1.0\rightarrow0.1\); RL weight \(\rho:1.0\rightarrow2.0\), over 80 updates; turn-level disagreement--uncertainty reweighting enabled \\
ECOD & Eight teacher rollouts per task condition; current-batch student outcomes; Beta(1,1) competence priors; empirical survival of successful teacher lengths \\
\bottomrule
\end{longtable}

\hypertarget{appendix-evaluation-details}{%
\subsection{Evaluation details}\label{appendix-evaluation-details}}

\label{sec:evaluation}

Evaluation used the student alone, without teacher intervention, at a sampling temperature of 0.4 and the full interaction horizon of each benchmark. Our primary metric was \textbf{success rate}, the percentage of episodes satisfying the environment's task-completion criterion. On ALFWorld, we reported results separately for the full seen and unseen evaluation sets. On WebShop, we additionally reported \textbf{task score}, the mean normalized reward expressed as a percentage of its maximum value of 1, to quantify partial satisfaction of the shopping requirements.

\textbf{Reported checkpoints.} Main results use the designated checkpoint for each configuration; auxiliary metrics use the same checkpoint. Ablations use update 150.

Repeated evaluations kept the task sequence fixed and varied the decoding seed. For Qwen2.5-1.5B-Instruct, the core RL/OPD/ECOD comparison used eight decoding runs for each ALFWorld split. The SOD, TCOD-F2B, and ATOD comparisons used one run on ALFWorld seen and eight on unseen; all reported WebShop methods used eight runs on the same 128-task schedule. Repeated runs are summarized by their mean and sample standard deviation. Main tables omit the seen standard deviation: the six core 1.5B RL/OPD/ECOD configurations have eight identical seen scores, whereas the remaining seen entries are single-run scores. The Qwen2.5-7B-Instruct experiments on ALFWorld used one seen run and eight unseen runs per evaluated model. All runs used training seed 0; the repetitions quantify decoding variability at fixed weights, not variability across independent training runs.

The abstract averages paired relative success-rate changes across SOD, TCOD-F2B, and ATOD with both student models, giving six equally weighted comparisons per benchmark. Each change is 100 times the ECOD--baseline difference divided by the baseline, computed from unrounded means and retaining negative values. ALFWorld uses unseen results; the two benchmarks are averaged separately.

\hypertarget{appendix-matched-allocation-ablations}{%
\subsection{Matched-allocation ablations}\label{appendix-matched-allocation-ablations}}

\label{sec:app-ablation}

The WebShop component ablations used Qwen2.5-1.5B-Instruct with GRPO, eight H200 GPUs, and 150 training updates. They shared the initialization, frozen teacher, task schedules, and optimization settings of the full ECOD reference. The teacher-KL coefficient remained 0.02. GRPO, OPD, and full ECOD results were reused from the matched main-experiment protocol; the three factor-removal controls and shared-prior control were trained separately.

Let \(i\) index response-turn samples in a control's current rollout batch \(X\), \(n_i\) denote the number of valid response tokens, and \(g_i=C_iF_iH_{T,i}\) denote the full ECOD weight evaluated on that same batch. For a control weight \(b_i\), we used

\begin{equation}
D(X)=\sum_i n_i g_i,\qquad
\widetilde b_i=
\begin{cases}
\displaystyle b_i\frac{D(X)}{\sum_j n_j b_j}, & D(X)>0,\\[6pt]
0, & D(X)=0.
\end{cases}
\label{eq:matched-allocation}
\end{equation}

The unscaled controls were \(b_i=F_iH_{T,i}\) (w/o competence weighting), \(b_i=C_iH_{T,i}\) (w/o outcome selection), \(b_i=C_iF_i\) (w/o temporal weighting), and \(b_i=C_iF_i\overline H(t_i)\) (pooled temporal prior). The shared curve was computed by pooling all 11,244 successful teacher trajectories from the 2,400 training task-condition entries, with equal weight per trajectory:

\begin{equation}
\overline H(t)=\frac{1}{11244}\sum_{\ell=1}^{11244}\mathbf 1[L_\ell>t].
\label{eq:shared-prior}
\end{equation}

\textbf{Matching protocol.} The prior uses training teacher trajectories, and the loss denominator counts all valid response tokens. Each control matches the batch-level weighted-token budget of full ECOD on the same rollouts, while redistributing that budget according to the ablated allocation rule. Matching concerns this total, not realized KL, gradient magnitude, or token count.

\textbf{Numerical checks.} Across the completed runs, the maximum relative mismatch in token-weighted supervision totals was below \(8\times10^{-16}\). A positive target with zero control support is treated as an error.

\hypertarget{appendix-teacher-strength-and-allocation-controls}{%
\subsection{Teacher-strength and allocation controls}\label{appendix-teacher-strength-and-allocation-controls}}

\label{sec:teacher-strength-details}

We used ALFWorld with Qwen2.5-1.5B-Instruct and GRPO, retaining the initialization, frozen teacher, training task sequence, and optimization settings of the corresponding main experiment. Every configuration was evaluated at the fixed training endpoint, step 150, on the same 134 unseen tasks with temperature 0.4 and no teacher intervention. Table A4 reports means and sample standard deviations over decoding seeds 0--7 for one training run per configuration. GRPO, OPD, and full ECOD reuse their existing endpoint evaluations.

\textbf{Strength controls.} OPD applies a uniform teacher coefficient of 0.02 or 0.002. Linear annealing uses \(\lambda(s)=0.004(150-s)/149\) for updates \(s=1,\ldots,150\); its update-average coefficient is 0.002, matching the smaller constant, without equalizing token counts or cumulative KL. All three coefficient schedules were prespecified and are reported. The comparison used mean Seen validation success over updates 130, 135, 140, 145, and 150. The constant 0.002 control ranked highest (76.56\%) and was selected, followed by linear annealing (76.41\%) and the constant 0.02 control (76.25\%).

\textbf{Allocation controls.} Uniform matching sets \(b_i=1\) in Eq.~\eqref{eq:matched-allocation}, giving every valid response token the same weight \(D(X)/\sum_i n_i\), with base teacher coefficient 0.02. The target \(D(X)\) is computed using full ECOD on the control's own current rollouts. The maximum relative matching error over 150 updates was \(9.73\times10^{-15}\); 147 batches changed allocation and three had zero target budget. The additional control without competence weighting uses \(b_i=F_iH_{T,i}\) under the same matching rule.

\textbf{Frozen-budget control.} Frozen-budget uniform allocation replays the weighted-token budget \(D_m^{\mathrm{full}}\) recorded at each update \(m\) of the full ECOD run. Every valid response token in the control batch receives weight \(D_m^{\mathrm{full}}/N_m^{\mathrm{ctrl}}\), where \(N_m^{\mathrm{ctrl}}\) is its valid response-token count; the base teacher coefficient remains 0.02. All 150 update budgets are matched, with a maximum relative allocation error of \(4.64\times10^{-15}\) before actor-side floating-point conversion. This control preserves the reference budget sequence while replacing selective allocation with uniform weighting.

\textbf{Table A4. ALFWorld teacher-strength and allocation controls (1.5B, GRPO).} Fixed step 150; mean ± sample s.d. across eight decoding runs.

\begin{longtable}[]{@{}lr@{}}
\toprule
Configuration & Unseen success (\%) \\
\midrule
\endhead
GRPO & 57.56 ± 1.23 \\
OPD, fixed 0.02 & 71.27 ± 2.22 \\
OPD, fixed 0.002 & 72.20 ± 2.70 \\
OPD, linear 0.004 to 0 & 67.35 ± 2.79 \\
Current-batch uniform & 64.93 ± 0.00 \\
Frozen-budget uniform & 72.29 ± 3.15 \\
w/o competence weighting & 71.46 ± 0.77 \\
ECOD & \textbf{74.16 ± 0.97} \\
\bottomrule
\end{longtable}

\hypertarget{appendix-learning-dynamics-protocol}{%
\subsection{Learning-dynamics protocol}\label{appendix-learning-dynamics-protocol}}

\label{sec:dynamics-protocol}

Figure~\ref{fig:dynamics} uses unsmoothed records from one training run per configuration. Shading covers updates 126--150; annotations report the relative change between the ECOD and baseline means over the five validation events in that interval. Teacher coefficients include each method's native weighting or annealing. SOD and ATOD use response-row means, whereas TCOD-F2B uses token-weighted means. Teacher coefficients are reported on method-specific scales and compared within each method pair.

\subsection{Supervision coverage and component distributions}
\label{sec:gate-diagnostics}

To characterize how ECOD allocates supervision beyond average teacher coefficients, we reconstruct its components from GRPO-based training rollouts and cached teacher records on ALFWorld and WebShop with Qwen2.5-1.5B-Instruct and Qwen2.5-7B-Instruct. Each setting includes one training run, all 150 updates, 2,400 task conditions, and 19,200 trajectories.

\textbf{Measurement.} Active-turn coverage is the fraction of recorded action turns assigned a positive ECOD weight. We weight task conditions equally for \(q\), \(\mu_T-\mu_S\), and \(C\), trajectories equally for \(F\), and active turns equally for \(H_T\) and \(g\). All distributions retain zeros, except the explicitly conditional nonzero-weight summary.

\begin{figure}[!htbp]
\centering
\includegraphics[width=\linewidth]{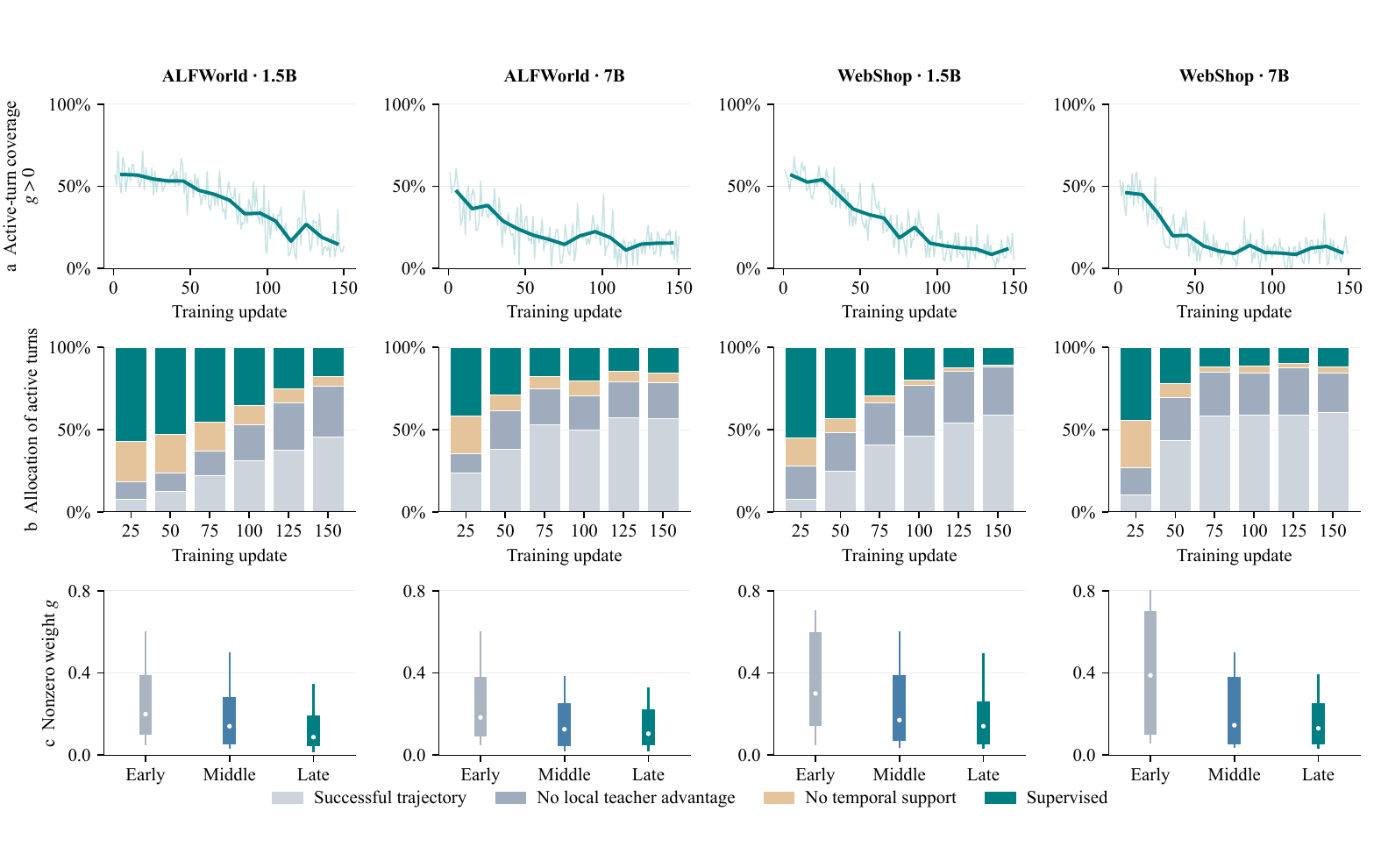}
\caption{\small \textbf{Supervision allocation across training.} Rows show active-turn coverage, its ordered outcome/competence/temporal partition, and nonzero weights. Curves show raw updates and ten-update pools; bars pool 25 updates. Weight summaries show median, interquartile, and 10--90\% ranges for early/middle/late stages (50 updates each).}
\label{fig:gate-coverage}
\end{figure}

\begin{figure}[!htbp]
\centering
\includegraphics[width=\linewidth]{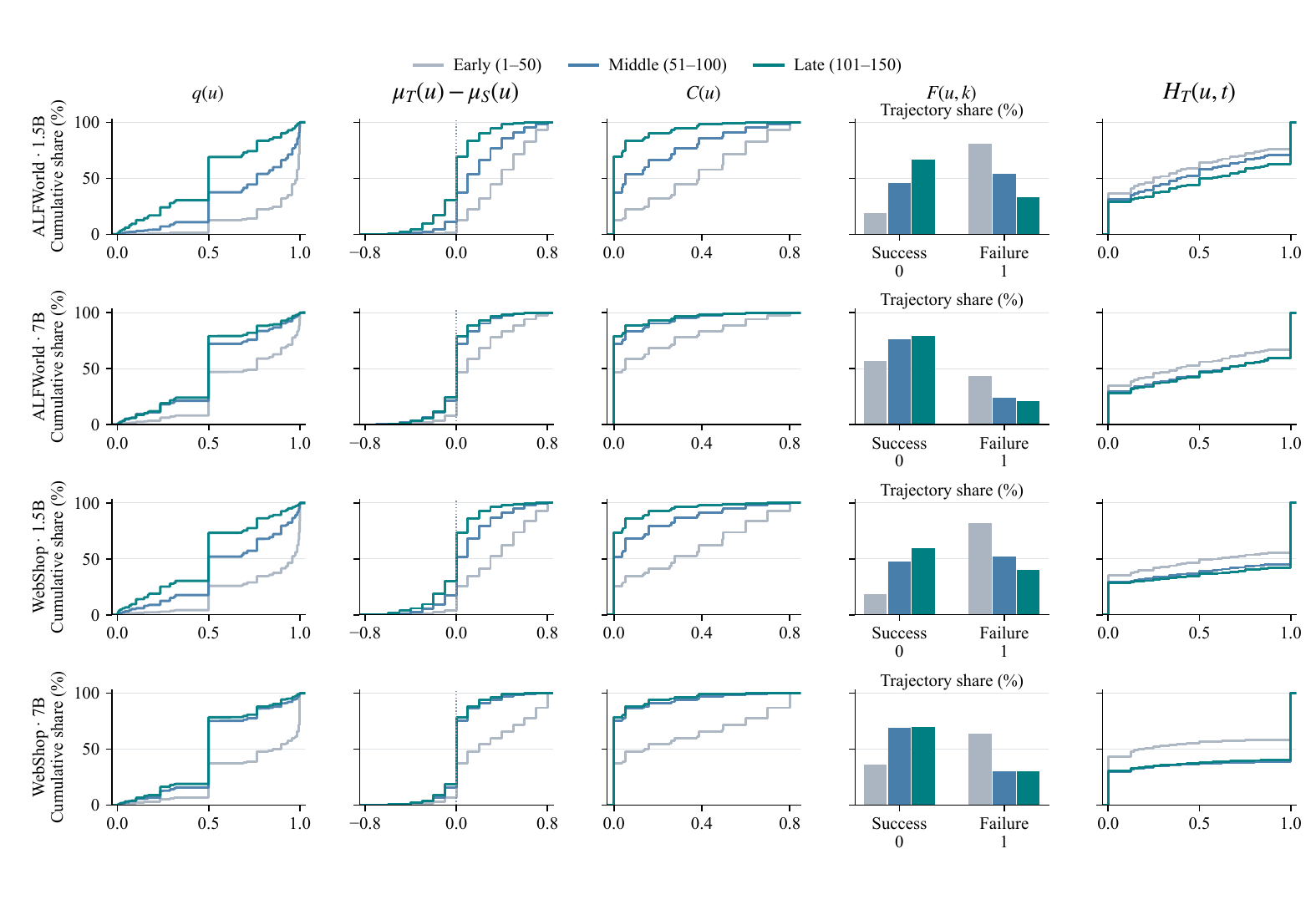}
\caption{\small \textbf{Component distributions across training stages.} Curves show empirical cumulative distributions from one training run per setting; bars show the binary distribution of \(F\). Gray/blue/teal denote updates 1--50/51--100/101--150. Each stage contains 800 task conditions and 6,400 trajectories per setting.}
\label{fig:gate-components}
\end{figure}

\textbf{Observed allocation.} Across all updates, nonzero weights cover 41.70\% and 24.40\% of active turns on ALFWorld, and 30.66\% and 21.35\% on WebShop, for the 1.5B and 7B students, respectively (Figure~\ref{fig:gate-coverage}). Figure~\ref{fig:gate-components} reports the corresponding component distributions. Figure A2 shows how the competence, outcome, and temporal factors evolve across training stages.

%% file: sections/motivation_protocol.tex
\subsection{Motivation analysis protocol}\label{sec:motivation-protocol}

Figure~\ref{fig:motivation} uses Qwen2.5-1.5B-Instruct students in synthetic WebShop with 1,000 products and a 15-turn limit. The fixed teacher is the corresponding GRPO policy after 150 updates.

\textbf{Task and turn panels.} Panels (a) and (c) use one GRPO-based OPD run over updates 1--150. Each update contains 16 task conditions, with eight student rollouts and eight cached teacher rollouts per condition. The 2,400 entries are task-condition observations across training and may include repeated tasks. Panel (c) groups entries by teacher and student success counts. Among groups with at least four teacher successes and more teacher than student successes, it shows the four largest groups, containing 132, 59, 49, and 35 conditions. In the 8:7 group, each point is the mean of eight successful teacher lengths for one condition; these means range from 4.5 to 10.5 turns. Heatmap columns show condition-specific length distributions normalized by the number of successful teacher trajectories.

\textbf{Action-score panel.} Panel (b) pools successful trajectories from GRPO, OPD, and SOD at checkpoint 125, using rollout update 126, and TCOD-F2B at checkpoint 145, using rollout update 146. The four 128-trajectory batches contain 303 successful trajectories. A fixed SHA-256 rule selects one action turn per trajectory. Two OPD responses without a unique nonempty action span are excluded without replacement. The remaining 301 actions comprise 73 GRPO, 69 OPD, 71 SOD, and 88 TCOD-F2B samples.

The generating student checkpoint and the fixed teacher score identical action tokens under the recorded history and response prefix, including preceding reasoning. Scores are mean log probabilities over action-content tokens. The scatterplot shows exponentiated scores, and the histogram shows student-minus-teacher differences in nats per token. Each action has equal weight.

%% file: sections/recent_opd_comparison.tex
\subsection{Comparison with recent OPD methods}
\label{sec:recent-opd-comparison}

Table~\ref{tab:recent-opd-comparison} distinguishes the evidence used for supervision, the affected update term, and changes to online student rollouts.

\input{tables/recent_opd_comparison}

RetireOPD uses global teacher-student completion rates to determine teacher retirement. ECOD recomputes task-conditioned completion statistics and weights individual failed trajectories. TurnOPD uses successful student lengths and KL statistics to control rollout depth; ECOD uses successful teacher lengths to weight turns within retained student rollouts. OVCSD and HDPO select all-failure groups and verify teacher continuations or solutions. ECOD's trajectory gate also selects failures within mixed-outcome groups. Table~\ref{tab:recent-opd-comparison} records the evidence and intervention locations of the other methods.

Related analyses examine token-support matching \citep{revisitingopd}, teacher-student compatibility and prompt selection \citep{rethinkingopd}, and the broader design space of OPD objectives and training dynamics \citep{opdsurvey}.

%% file: tables/recent_opd_comparison.tex
\begin{table}[H]
\centering
\caption{Comparison of supervision evidence and intervention mechanisms.}
\label{tab:recent-opd-comparison}
\begingroup
\scriptsize
\setlength{\tabcolsep}{2pt}
\renewcommand{\arraystretch}{1.12}
\newcommand{\opdyes}{\textcolor[rgb]{0.00,0.36,0.20}{\ensuremath{\checkmark}}}
\newcommand{\opdpartial}{\textcolor{gray}{\ensuremath{\triangle}}}
\newcommand{\opdno}{\textendash}
\begin{tabular*}{\linewidth}{@{\extracolsep{\fill}}l*{5}{c}@{}}
\toprule
& \multicolumn{3}{c}{Supervision evidence} & Optimization & Sampling \\
\cmidrule(lr){2-4}\cmidrule(lr){5-5}\cmidrule(l){6-6}
Method & \shortstack{Completion evidence\\and scope} & Outcome & Turn & Teacher term & \shortstack{Online rollout\\retained} \\
\midrule
\multicolumn{6}{l}{\textit{Direct experimental comparators}} \\
SOD~\citep{sod} & \opdno & \opdno & \opdyes & Loss & \opdyes \\
TCOD-F2B~\citep{tcod} & \opdno & \opdno & \opdpartial & Loss & \opdno \\
ATOD~\citep{atod} & \opdno & \opdno & \opdyes & Teacher advantage & \opdyes \\
\midrule
\multicolumn{6}{l}{\textit{Adaptive strength and gating}} \\
RetireOPD~\citep{retireopd} & Global T--S rates & \opdpartial & \opdno & Loss & \opdyes \\
SDAR~\citep{sdar} & \opdno & \opdpartial & \opdno & Loss & \opdyes \\
SEED~\citep{seedopd} & Completed trajectories & \opdpartial & \opdno & Loss & \opdyes \\
SAF-OPD~\citep{safopd} & \opdno & \opdno & \opdno & Teacher advantage & \opdyes \\
SERL~\citep{serl} & Reward-advantage sign & \opdpartial & \opdyes & Loss + RL advantage & \opdyes \\
\midrule
\multicolumn{6}{l}{\textit{Turn and temporal allocation}} \\
Guided-OPD~\citep{guidedopd} & \opdno & \opdno & \opdpartial & Loss & \opdno \\
TurnOPD~\citep{turnopd} & Student probe successes & \opdpartial & \opdyes & Loss & \opdno \\
ReOPD~\citep{reopd} & \opdno & \opdno & \opdyes & Loss & \opdno \\
STRIDE~\citep{stride} & \opdno & \opdno & \opdyes & Loss & \opdno \\
\midrule
\multicolumn{6}{l}{\textit{Outcome selection and future validation}} \\
FutureBridge-OPD~\citep{futurebridge} & \opdno & \opdpartial & \opdpartial & Teacher advantage & \opdno \\
OVCSD~\citep{ovcsd} & Group + repair outcomes & \opdyes & \opdpartial & Loss + comparative RL & \opdno \\
HDPO~\citep{hdpo} & Group + solution outcomes & \opdyes & \opdno & Loss & \opdno \\
RLRT~\citep{rlrt} & Student correctness & \opdyes & \opdno & Reward advantage & \opdyes \\
\midrule
\multicolumn{6}{l}{\textit{Peer, skill, and context construction}} \\
MOPD~\citep{mopd} & Peer rollout outcomes & \opdyes & \opdno & Loss & \opdyes \\
OPID~\citep{opid} & Completed trajectories & \opdpartial & \opdpartial & Teacher advantage & \opdyes \\
CCOPD~\citep{ccopd} & \opdno & \opdno & \opdpartial & Loss & \opdyes \\
\midrule
\textbf{ECOD (ours)} & Task-conditioned T--S counts & \opdyes & \opdyes & Loss or teacher advantage & \opdyes \\
\bottomrule
\end{tabular*}
\par
\raggedright
\opdyes\ explicit mechanism; \opdpartial\ related mechanism with different granularity; \opdno\ absent from the allocation or target-construction rule. Completion evidence excludes ordinary RL rewards and specifies how outcomes enter the teacher-related rule (T--S: teacher--student). Outcome denotes trajectory selection; turn denotes within-trajectory allocation. The teacher-term column follows original formulations. SOD specifies a weighted OPD loss; our shared-framework implementation applies its step-weighting mechanism through a teacher-derived advantage. Online rollout retained refers to the base online student rollout procedure. ECOD separately precollects teacher trajectories for allocation statistics.\par
\endgroup
\end{table}